\documentclass[conference]{IEEEtran}
\IEEEoverridecommandlockouts
\usepackage{cite}
\usepackage{amsmath,amssymb,amsfonts}
\usepackage{amsthm}
\usepackage{algorithmic}
\usepackage{graphicx}
\usepackage{caption}
\usepackage{textcomp}
\usepackage{xcolor}
\usepackage{float}

\begin{document}

\title{FROC: A Unified Framework with Risk-Optimized Control for Machine Unlearning in LLMs
}


\author{
Si Qi Goh$^{\dagger\ddagger}$,
Yongsen Zheng$^{\dagger\ddagger}$,
Ziyao Liu$^{\dagger}$,
Sami Hormi$^{\dagger}$,
and Kwok-Yan Lam$^{\dagger\ddagger\star}$\\[6pt]
$^{\dagger}$Digital Trust Centre, Nanyang Technological University, Singapore\\
$^{\ddagger}$College of Computing and Data Science, Nanyang Technological University, Singapore\\
\small{
siqi005@e.ntu.edu.sg,\;
yongsen.zheng@ntu.edu.sg,\;
liuziyao@ntu.edu.sg,\;
sami.hormi@ntu.edu.sg,\;
kwokyan.lam@ntu.edu.sg$\star$
}
}


\maketitle

\begin{abstract}
Machine unlearning (MU) seeks to eliminate the influence of specific training examples from deployed models. As large language models (LLMs) become widely used, managing risks arising from insufficient forgetting or utility loss is increasingly crucial. Current MU techniques lack effective mechanisms for evaluating and controlling these risks, hindering the selection of strategies that appropriately balance safety and utility, and raising trust concerns surrounding the "right to be forgotten." To address these issues, we propose FROC, a unified framework with Risk-Optimized Control for machine unlearning in LLMs. FROC is built around a conformal-style risk-control formulation that expresses a user-specified risk budget on unlearning behavior. This probability-based constraint enables FROC to compare MU strategies, identify feasible operating regions, and guide hyperparameter selection according to desired trade-offs between forgetting sufficiency and utility preservation. To operationalize this constraint, FROC introduces a smoothly varying continuous risk model that aggregates forgetting deficiency and utility degradation into a single configuration-level score. Building on conformal risk analysis, FROC computes (1) the Conformal Unlearning Risk (CUR), a data-driven estimated value on the probability that forgotten samples continue to influence model predictions, and (2) risk-controlled configuration sets, which identify unlearning hyperparameters that are valid under the specified risk budget. Experiments across multiple LLM MU methods demonstrate that FROC produces stable, interpretable risk landscapes and reveals consistent relationships between unlearning configurations, semantic shift, and utility impact. FROC reframes MU as a controllable, risk-aware process and offers a practical foundation for managing unlearning behavior in large-scale LLM deployments.

\end{abstract}


\section{Introduction}

The rapid adoption of large-scale machine learning systems, particularly foundation models, graph-based models \cite{HiCore}, and large language models (LLMs) has intensified longstanding concerns around data privacy, individual rights, and model accountability. These models are known to memorize rare or unique samples, inadvertently storing personal information, copyrighted text, or other sensitive artifacts. In real-world applications, data owners may demand that their data be removed from a trained language model due to privacy or copyright concerns, as mandated, for example, by the General Data Protection Regulation \cite{voigt2017eu}. As a result, deployed AI systems now face growing legal and societal expectations to provide mechanisms that allow users to retract their data and ensure that models no longer retain or exploit the forgotten information. This challenge has given rise to the field of machine unlearning (MU), refers to the process of selectively removing specific training data points and their influence on an trained model, making the updated model behave the same as a model that was never trained on that data \cite{nguyen2025survey}.

Existing MU approaches ranging from gradient-ascent removal \cite{neel2021descent} and targeted model editing \cite{yao2023editing} to SISA \cite{golatkar2020eternal} and distillation typically produce highly diverse outcomes. They typically demonstrate their success of removal through behavioral metrics \cite{ginart2019making}, such as degraded performance on forgotten samples or reduced membership inference risk. However, the problem is not merely whether unlearning is performed, but whether it is performed in a way that appropriately manages the associated risks.

A key gap is the absence of a principled mechanism for \emph{controlling} these risks across different unlearning configurations \cite{neel2021descent, izzo2021approximate}. Practitioners require tools
to express conditions such as ensuring that most forgotten samples undergo a sufficient predictive shift, or utility losses remain within acceptable limits. To address this, we adopt a conformal-style, probability-based formulation of risk control which specifies a user-defined \emph{risk budget} on unlearning behavior. Instead of serving as a guarantee, this expression encodes a desired operating condition: an MU configuration is preferable when the proportion of samples exceeding a target risk level $\alpha$ is small, controlled by a tolerance parameter $\delta$.

We introduce \textbf{FROC}, a unified \emph{Framework with Risk-Optimized Control} in machine unlearning for LLMs. FROC operationalizes the above risk-control perspective by \textbf{(i) defining a continuous, softplus-smoothed unlearning risk function} that jointly quantifies insufficient forgetting and excessive utility degradation, and \textbf{(ii) aggregating these signals into configuration-level risk assessments}. This enables FROC to map risk landscapes, compare diverse MU strategies, and select configurations that satisfy either application or regulation-driven risk budgets. By reframing MU as a risk-management task, FROC provides a method-agnostic approach for understanding and steering unlearning behavior in large-scale LLMs.

\section{Related Work}


Early work framed MU as a data-deletion problem, relying on exact retraining methods \cite{ginart2019making}, but were limited to small-scale models. Later research introduced SISA for efficient unlearning through sharded pipelines \cite{golatkar2020eternal}. Other methods utilize structured retraining, influence function approximations, and noise injection and ensemble models \cite{foster2024fast, ye2025enhancing}. MU is also being incorporated with federated unlearning (FU) in various applications to ensure privacy \cite{liu2024guaranteeing}. Some of these works enables client-level data removal \cite{liu2021federaser, liu2025privacy} without complete retraining. While these techniques are useful, they exhibit variable behaviors and lack a unified framework to manage risks associated with unlearning configurations.


LLM-specific MU methods include targeted removal \cite{tarun2023fast}, distillation-based forgetting \cite{kim2024layer}, and parameter-editing techniques \cite{tarun2023fast}. These methods often have unstable forgetting–utility trade-offs and are sensitive to configuration choices despite having pratical utility. Research on memorization and data influence, such as extraction attacks \cite{liu2025threats}, indicates that residual memorization can persist even with high performance. This highlights the need for MU frameworks to quantify insufficient forgetting and degradation in a structured manner.

Conformal prediction has emerged as an influential framework for controlling the frequency of undesirable outcomes in machine learning. Beyond its original role in uncertainty quantification, conformal prediction has been extended to applications involving risk-sensitive or distribution-shift–aware decision making, such as Conformal Risk Control \cite{stankeviciute2021conformal}, predictive inference with Jackknife+, conditional-coverage methods, and calibration under covariate shift. These approaches share the goal of constraining how often a model exhibits undesirable behavior, whether in the form of miscalibrated predictions, excessive risk, or violations of fairness or safety criteria. Conceptually, this perspective is closely related to machine unlearning: MU also seeks to regulate the frequency of adverse behaviors. More specifically, instances where the model insufficiently forgets targeted data or excessively degrades utility on retained samples. Conformal-style probabilistic control therefore provides a natural foundation for thinking about unlearning as a risk-management problem.

\section{Preliminaries}


This section presents: (i) a surrogate measure of forgetting strength, (ii) a utility-degradation metrics, and (iii) conformal-style risk quantity. These are crucial for evaluating unlearning configurations in the conformal prediction framework.




\subsection{Unified Continuous Unlearning Risk Function}

\paragraph{\textbf{Forgetting deficiency}}

Let $\theta$ denote the original model and $\theta'$ the model obtained after
applying an unlearning configuration~$\lambda$. Let $D_{\mathrm{forget}}$ be the
forget set. We quantify the extent to which configuration~$\lambda$ suppresses the
forget set via the surrogate forgetting-shift score
\begin{equation}
\begin{aligned}
s^{(\lambda)}
=\;& \log\!\bigl(\mathrm{Loss}_{U}(\lambda)\bigr)
\\
&+ \bigl(\max_{\lambda'} \mathrm{Acc}_{U}(\lambda') - \mathrm{Acc}_{U}(\lambda)\bigr),
\end{aligned}
\label{eq:forget-shift}
\end{equation}
where $\mathrm{Loss}_{U}(\lambda)$ denotes the average forget-set loss (equivalently, $\log\mathrm{PPL}$), and $\mathrm{Acc}_{U}(\lambda)$ denotes the forget-set accuracy.
Larger values of $s^{(\lambda)}$ correspond to stronger forgetting effects.

\paragraph{\textbf{Utility Degradation}}

To quantify changes induced on the retain distribution, we measure both shifts in loss and accuracy. Let $\mathrm{Loss}_{R}(\lambda)$ and $\mathrm{Acc}_{R}(\lambda)$ denote
the retain-set loss and accuracy under configuration $\lambda$. We define
\begin{equation}
\begin{aligned}
r^{(\lambda)} =\;& 
\log\!\bigl(\mathrm{Loss}_{R}(\lambda)\bigr)
-
\log\!\left(\min_{c'} \mathrm{Loss}_{R}(\lambda')\right)
\\
&+
\bigl(\max_{\lambda'} \mathrm{Acc}_{R}(\lambda') - \mathrm{Acc}_{R}(\lambda)\bigr),
\end{aligned}
\label{eq:retain-distortion}
\end{equation}
which measures distortion relative to the best-performing configuration. 
Any increase in retain-set distortion contributes directly to the unified risk,  and the conformal adjustment subsequently determines which levels of degradation  are statistically acceptable under the user's risk budget~$\delta$.

To support a smooth evaluation criterion, we define the following soft margins:
\begin{align}
\Delta_f(\lambda)
&=
\mathrm{softplus}\!\bigl(
\hat{\alpha}_{\mathrm{unlearn}}
-
s^{(\lambda)}
\bigr),
\label{eq:forget-penalty}\\
\Delta_u(\lambda)
&=
\mathrm{softplus}\!\bigl(
r^{(\lambda)}),
\label{eq:utility-penalty}
\end{align}
where $\mathrm{softplus}(z) = \log(1 + e^{z})$.

These quantities are used in later sections to construct a unified, risk-oriented evaluation of unlearning configurations.

\paragraph{\textbf{Unified continuous per-configuration risk}}
To avoid cancellation effects and to permit prioritization between forgetting sufficiency and utility preservation, we combine the two penalties using non-negative weights $w_f, w_u$:
\[
\widetilde{R}(\lambda)
\;=\;
w_f\, \Delta_f(\lambda) \;+\; w_u\, \Delta_u(\lambda).
\]
This unified risk increases monotonically whenever forgetting sufficiency worsens or utility degradation grows. For experimental purposes, we set $w_f = w_u = 1$, but these weights can be tuned to reflect a user's preference for prioritizing stronger forgetting or  stricter utility preservation.

\paragraph{\textbf{Aggregate unlearning risk for control}}
The overall risk for an unlearning configuration is obtained by averaging \(\widetilde{R}(\lambda)\) over the evaluation dataset, yielding a scalar violation score that supports principled control and selection.

\subsection{Conformal-Style Risk Control}

Let $\hat{D}_{\mathrm{ref}}$ be a reference set disjoint from the forget set, and let $\hat{R}_{\theta'}(\hat{D}_{\mathrm{ref}})$ denote the unified unlearning risk of the post-unlearning model $\theta'$. Following the functional form of conformal prediction, we compute a scalar risk statistic
\begin{equation}
\mathbb{P}_{(x,y)\sim\mathcal{D}}
\!\left[
R(p_{\theta'}(x),y)
\le
\hat{\alpha}_{\mathrm{unlearn}}
\right]
\approx 1-\delta,
\end{equation}
with
\begin{equation}
\begin{aligned}
\hat{\alpha}_{\mathrm{unlearn}}
=\min\Bigl\{&
h^{-1}\!\left(\tfrac{\ln(1/\delta)}{N_{\mathrm{ref}}};
\hat{R}_{\theta'}(\hat{D}_{\mathrm{ref}})\right),
\\
&\Phi^{-1}_{\mathrm{bin}}\!\left(\tfrac{\delta}{e};
N_{\mathrm{ref}},\hat{R}_{\theta'}(\hat{D}_{\mathrm{ref}})\right)
\Bigr\},
\end{aligned}
\end{equation}
where $h(a,b)=a\log(a/b)+(1-a)\log\frac{1-a}{1-b}$ and $h^{-1}(\cdot;b)$ is its partial inverse. This quantity serves as a model-behavior statistic used throughout the paper.

\section{Risk Optimized Control Machine Unlearning via Conformal Analysis}
Section III introduced the forgetting-shift statistic, utility-degradation measure, and the risk metric \(\hat{\alpha}_{\mathrm{unlearn}}\). We now use these to develop a methodology for selecting unlearning configurations that balance forgetting with retain set preservation under a risk budget.

\subsection{Conformal Analysis Principle and FROC Framework}

Conformal analysis is crucial in FROC, providing a probabilistic mechanism to limit undesirable behaviors from unlearning configurations. An adverse event occurs when the risk function \(R(p_{\theta'}(x),y)\) exceeds a user-defined threshold \(\alpha\), indicating the maximum acceptable level of semantic retention or utility degradation post-unlearning. Using a reference dataset, we compute the controlled risk as the fraction of samples where the risk exceeds \(\alpha\), estimating how often model shows unsatisfactory behavior. Conformal reasoning allows us to view this quantity as a control signal: rather than treating the inequality \(\mathbb{P}[R \le \alpha] \ge 1-\delta\) as a formal guarantee, we interpret it as a \emph{risk-budget constraint} defining the tolerable rate \(\delta\) of unlearning violations.
Consequently, FROC repurposes conformal ideas, where we do not directly certify predictions, but to shape the \emph{risk landscape} of unlearning so that conformal assessment, probabilistic adjustment, and user-defined risk jointly guide the selection of MU configurations. Given $N_{ref}$ risk evaluations:
\begin{equation}
\hat{R} = \frac{1}{N_{ref}} \sum_{i=1}^{N_{ref}} R(p_{\theta'}(x_i), d),
\end{equation}
FROC defines:
\begin{equation}
\hat{\alpha}_{unl} = \min\left(
h^{-1}\!\left(\tfrac{\ln(1/\delta)}{N_{ref}}, \hat{R}\right),
\Phi^{-1}_{bin}\!\left(\tfrac{\delta}{e}; N_{ref},\hat{R}\right)
\right).
\end{equation}This yields:
\begin{equation}
P_{x\sim D_{ref}}\!\left[R(p_{\theta'}(x),d) \le \hat{\alpha}_{unl}\right] \ge 1-\delta.
\end{equation}

\begin{figure*}[t]
    \centering
    \includegraphics[width=\textwidth]{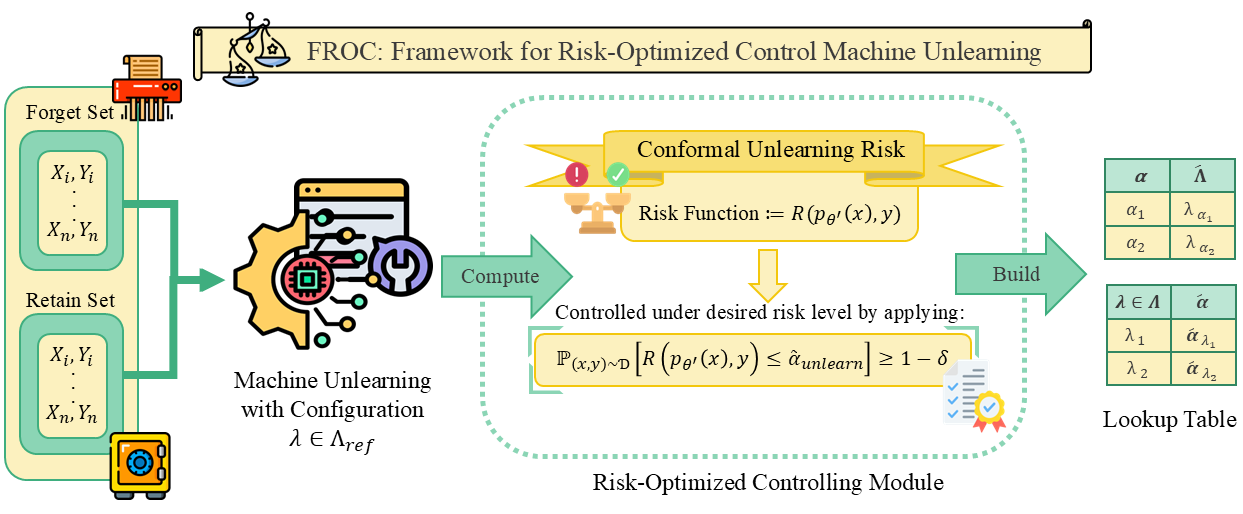}
    \caption{Overview of the FROC framework}
    \label{fig:froc-framework}
\end{figure*}
Figure~\ref{fig:froc-framework} illustrates the overall FROC framework, which organizes machine unlearning into a risk-optimized control pipeline. Given a designated forget set and retain set, an unlearning configuration $\lambda \in \Lambda_{\mathrm{ref}}$ is applied to a pre-trained model to obtain a perturbed model $\theta'$. The resulting model is evaluated through the conformal-style unlearning risk, which quantifies how frequently $\theta'$ exceeds a specified risk threshold on a held-out reference distribution. This statistic is then interpreted under a user-defined risk budget $\delta$, yielding an adjusted violation estimate that reflects the balance of the configurations $\lambda$ between effective forgetting and retain-set preservation. By comparing this adjusted risk against the allowable budget, the framework determines whether the configuration satisfies the desired operating criterion. Repeating this process across a grid of candidate configurations produces a lookup table mapping risk tolerances to admissible unlearning strengths, thereby enabling reliable, computationally efficient selection of unlearning parameters in downstream systems. The FROC framework thus unifies empirical risk assessment, conformal-style adjustment, and configuration selection into a control mechanism for stable unlearning behavior.

\begin{figure}[H]
\centering
\includegraphics[width=1.05\linewidth]{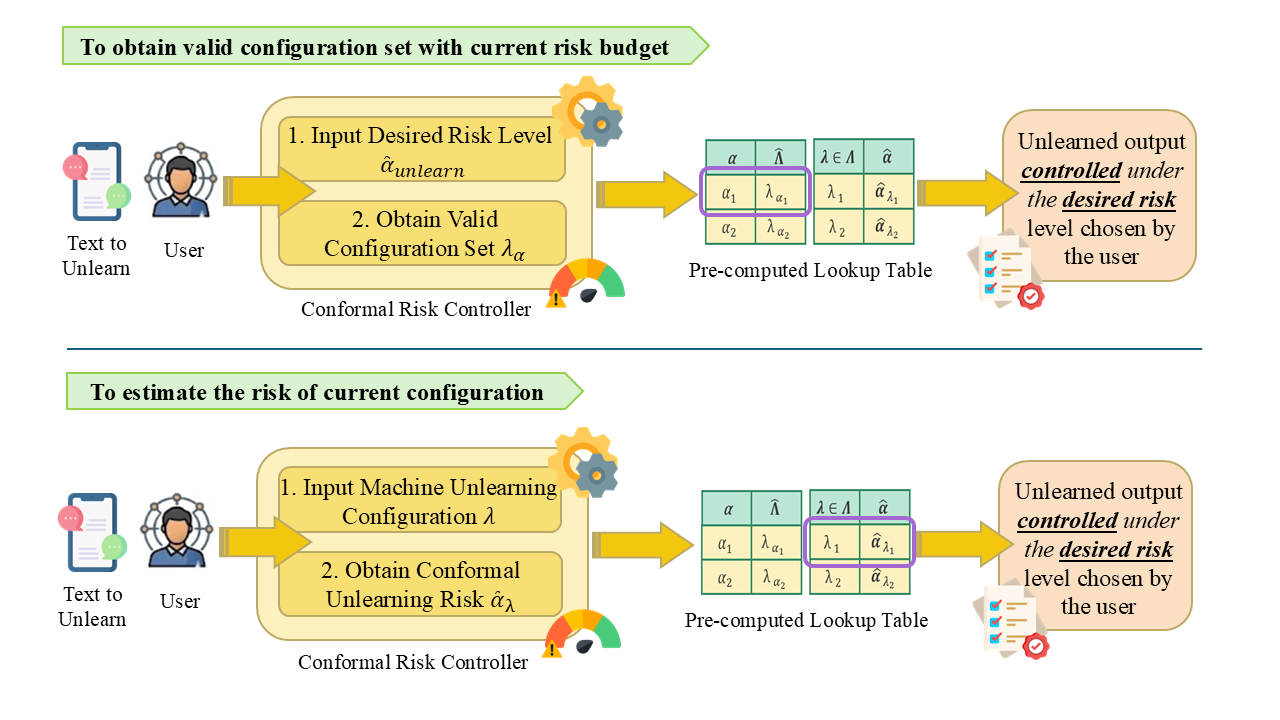}
\caption{How FROC works in the Inference Stage.}
\label{fig:froc-inference}
\end{figure}

After this pre-computing stage, the proposed FROC also include two-way controller. Figure \ref{fig:froc-inference} illustrates how the FROC controller is used in practice for both configuration selection and risk estimation. In the first mode, the user specifies a desired unlearning risk budget, and the controller queries the pre-computed lookup table, which is pre-constructed to retrieve the set of configurations whose adjusted risk estimates satisfy the specified threshold. This enables the system to select an unlearning configuration that meets the user’s tolerance while ensuring that the resulting model adheres to the desired threshold. In the second mode, the controller evaluates an input configuration by locating its corresponding entry in the lookup table and returning its controlled risk that was being computed by using the reference set. This provides users with an assessment of the expected risk associated with the chosen configuration.


\subsection{Unlearning Effectiveness Control}


We manage the unlearning risks associated with language models following different unlearning methods through conformal risk analysis. This methodology ensures that the conformal risks of the unlearned models are rigorously controlled, relying on test statistics derived from an in-distribution reference set. In this study, we consider a reference set that is representative of the target data distribution, allowing us to effectively evaluate the performance and reliability of the unlearned models. In this work, we consider a reference set:
\begin{equation}
\hat{\mathcal{D}}_{\text{ref}} = \left\{ (X_i, Y_i) \right\}_{i=1}^{N_{\text{ref}}}
\label{eq:ref-dataset}
\end{equation} with size \( N_{\text{ref}} \), and compute the unified risk of the unlearned model \( p_{\theta'} \) as:
\begin{equation}
\hat{R}_{\theta'}(\hat{\mathcal{D}}_{\text{ref}}) = \frac{1}{N_{\text{ref}}} \sum_{(x,y) \in \hat{\mathcal{D}}_{\text{ref}}} R(p_{\theta'}(x), y).
\label{eq:emp-ref-risk}
\end{equation}

\textbf{Condition 1: \textit{Probabilistic Risk-Control Condition for FROC}}

Let $D_{\text{forget}}$ be the forget set, and $\theta'$ the unlearned model obtained via unlearning under configuration $\lambda$. 

Let \( \hat{R}_{\theta'}(\hat{\mathcal{D}}_{\text{ref}}) \) be the empirical risk on a held-out reference set \( \hat{\mathcal{D}}_{\text{ref}} \), disjoint from the forget set. Then, with probability at least \( 1 - \delta \), the generalization risk of \( \theta' \) on a test sample, $(X_{\text{test}}, Y_{\text{test}}) \sim D$, satisfies:

\begin{equation}
\mathbb{P}_{(x, y) \sim \mathcal{D}} \left[ R(p_{\theta'}(x), y) \leq \hat{\alpha}_{\text{unlearn}} \right] \geq 1 - \delta,
\label{eq:prob-risk}
\end{equation}

\textit{
where the high-probability unlearning risk upper bound \( \hat{\alpha}_{\text{unlearn}} \), the so-called conformal unlearning risk, is given by:}

\begin{equation}
\begin{aligned}
\hat{\alpha}_{\text{unlearn}} = \min \bigg\{ & \,
h^{-1} \left( \frac{\ln(1/\delta)}{N_{\text{ref}}}; 
\hat{R}_{\theta'}(\hat{\mathcal{D}}_{\text{ref}}) \right), \\
& \Phi^{-1}_{\text{bin}} \left( \frac{\delta}{e}; N_{\text{ref}}, 
\hat{R}_{\theta'}(\hat{\mathcal{D}}_{\text{ref}}) \right)
\bigg\}
\end{aligned}
\label{eq:unlearn-risk-bound}
\end{equation}

\textit{
with \( h(a, b) = a \log(a/b) + (1 - a) \log \left( \frac{1 - a}{1 - b} \right) \), the partial inverse \( h^{-1}(h(a,b); a) = b \), and \( \Phi^{-1}_{\text{bin}} \) denoting the inverse binomial cumulative distribution function.
}

Condition 1 integrates the conformal-style statistic into the FROC framework by linking each unlearning configuration to its estimated controlled risk. As shown in Figures~\ref{fig:froc-framework} and \ref{fig:froc-inference}, the controlled risk $\hat{R}_{\theta'}(\hat{D}_{\mathrm{ref}})$ feeds into the conformal adjustment defined in Section~III, producing $\hat{\alpha}_{\mathrm{unlearn}}$. This value serves as a calibrated indicator of how frequently a configuration is expected to exceed the allowed risk threshold. Condition~1 therefore acts as the criterion by which the FROC controller determines whether a configuration is admissible under a user-specified risk budget. Evaluating this condition across all candidate configurations populates the lookup table used by the controller, allowing the system to retrieve valid configurations or assess the risk of a given configuration in real time. In this way, Condition~1 provides the operational bridge between the conformal-style metric and the practical control mechanisms that govern FROC's unlearning behavior.

\textbf{Condition 2: \textit{Optimal Configuration of Unlearning Algorithm}}


In the risk-optimized control framework (Proposition 1), we establish a desired risk level \( \alpha \) and compute a valid configuration set \( \hat{\Lambda}_\alpha \) for machine unlearning. This set is designed to ensure that the unlearning procedures applied within it result in post-unlearning risks that remain below the threshold \( \alpha \). To maintain rigorous control over the family-wise error rate, we employ the Bonferroni correction \cite{weisstein2004bonferroni}. We then assess empirical risks using randomly sampled test data, evaluating models that have been retrained with the configurations in \( \hat{\Lambda}_\alpha \). Our results demonstrate that the post-unlearning empirical risks consistently fall below the conformal unlearning risk bounds, validating the effectiveness of our approach in managing unlearning risks.

\vspace{0.5em}

\section{Evaluation}
\subsection{Experimental Setup}


We evaluate our risk-optimized unlearning framework on the RedPajamas \cite{weber2024redpajama} benchmark using a pretrained LLAMA3.1-8B \cite{touvron2023llama}. The dataset has over 1.2T tokens from various sources, and we use the subsets of it which contains over 73M tokens. A reference set \(\hat{D}_{\mathrm{ref}}\) computes the conformal-style unlearning risk. Unlearning methods include gradient-ascent (GA), GA plus Descent, and GA with KL divergence, tested across different configurations \(\lambda\). We report metrics on an evaluation set of 12,000 samples split evenly between forget and retain set, measuring forgetting effectiveness, utility preservation, and calibrated unlearning behavior. All models are trained with the AdamW optimizer at a learning rate of 0.00002 on NVIDIA A100-SXM4-40GB GPUs.

\subsection{Evaluation Results}

\begin{figure}[t]
    \centering
    \includegraphics[width=0.45\textwidth]{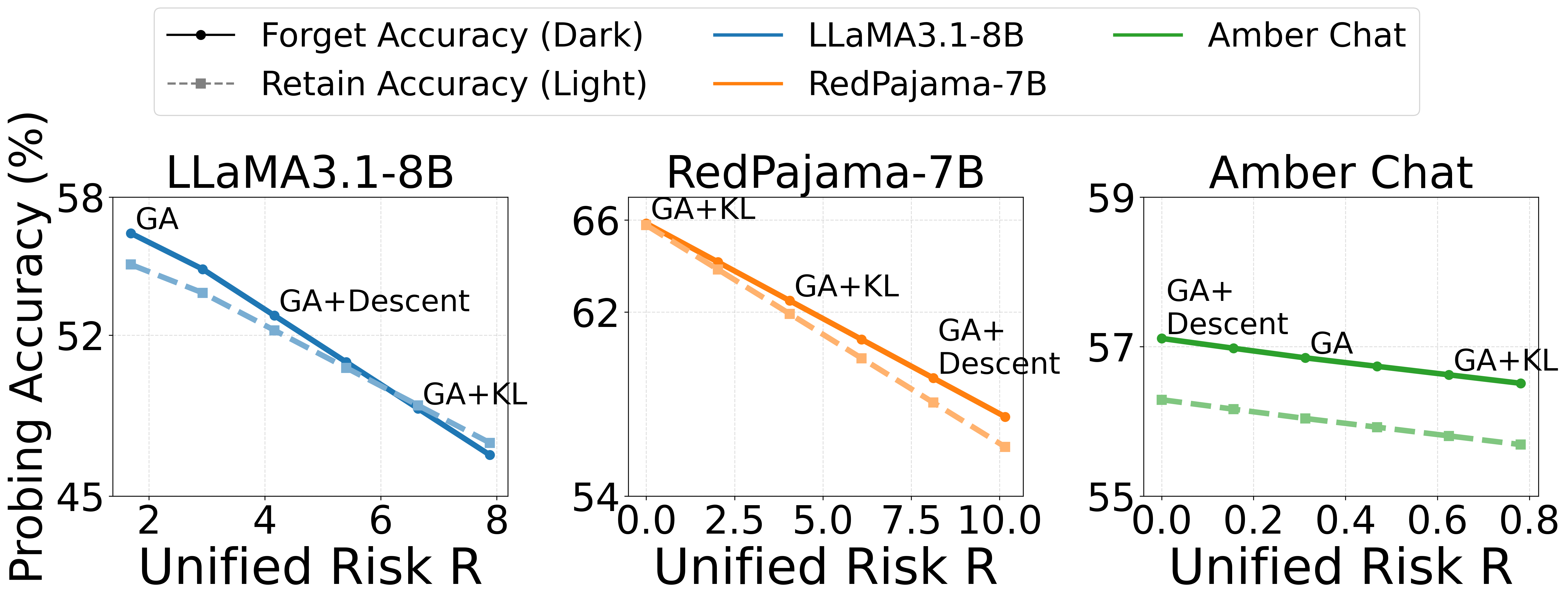}
    \caption{Unified risk versus probing accuracy across LLMs.}
    \label{fig:risk_acc_llms}
\end{figure}

To understand how unified risk $R(\lambda)$ governs the forgetting utility trade-off across different architectures, we begin by examining the probing accuracy of three LLMs under multiple unlearning configurations. Figure \ref{fig:risk_acc_llms} illustrates how the unified risk $R$ organizes the forgetting and utility trade-off across three LLMs under different unlearning configurations. For each model, the dark solid line tracks probing accuracy on forgotten samples, while the light dashed line tracks accuracy on retained samples. As unified risk increases, all models exhibit a consistent monotonic decline in both metrics, with forgotten-sample accuracy dropping more sharply, indicating stronger unlearning, while retained-sample accuracy degrades more gradually. The relative ordering of methods (GA, GA+Descent, GA+KL) aligns with their corresponding risk levels, demonstrating that low-risk configurations yield balanced forgetting and utility preservation, whereas high-risk configurations induce aggressive but unstable updates. Despite architectural and scale differences among LLaMA3.1-8B, RedPajama-7B, and Amber Chat, the overall trends remain structurally similar, showing that $R$ functions as a model-agnostic control variable that captures the stability and behavioral impact of unlearning across diverse LLM checkpoints.

\subsubsection{\textbf{Valid Configurations under Desired risk Level}}

\begin{figure}[t]
  \centering
  \includegraphics[width=0.8\linewidth]{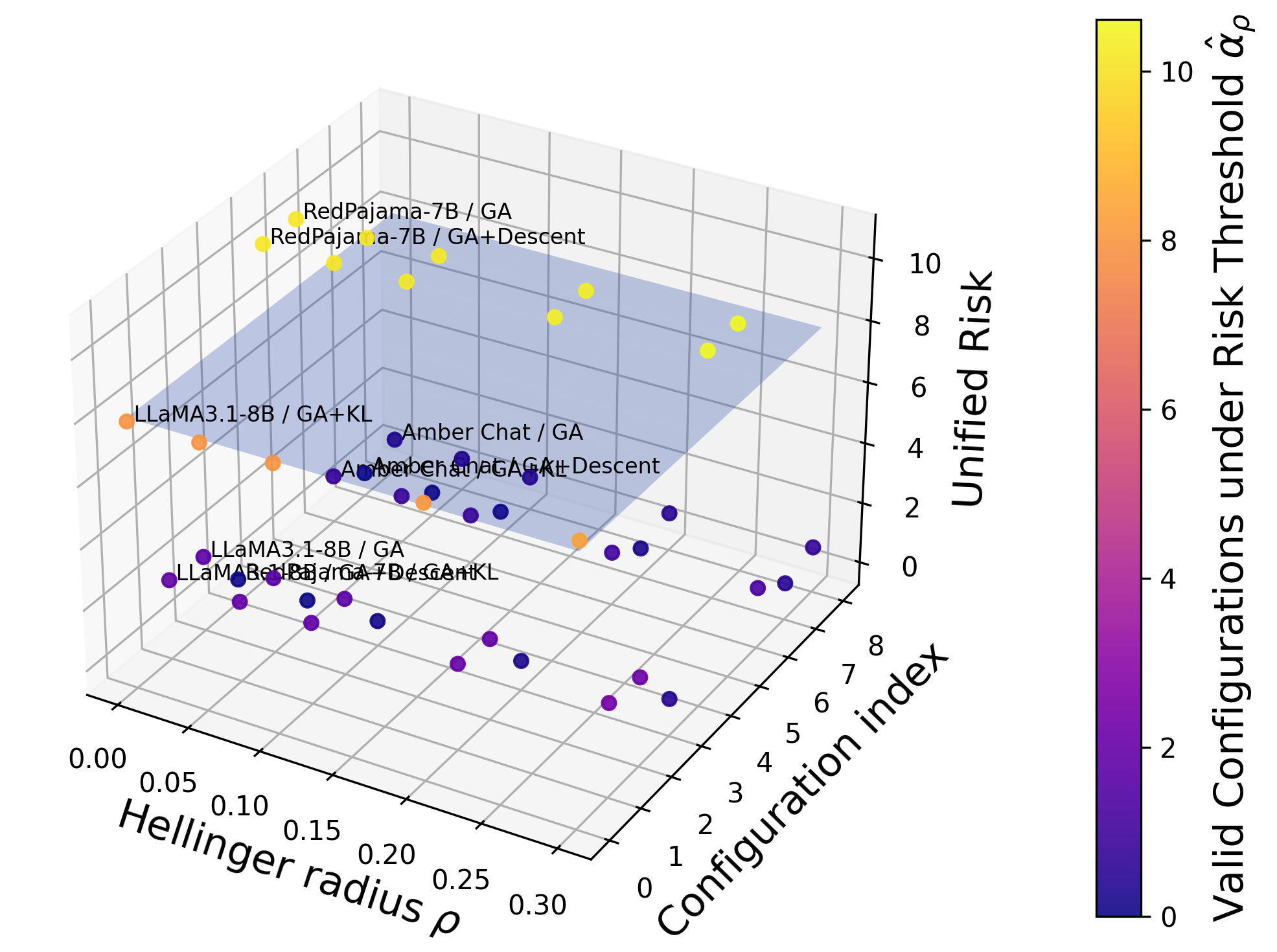}
  \caption{Valid configuration region under a target risk level $\alpha$. Points below the horizontal plane correspond to controlled-safe $(\lambda, N_{\mathrm{ref}})$ configurations.}
  \label{fig:3d_valid}
\end{figure} 

Figure \ref{fig:3d_valid} visualizes how unified risk behaves jointly across configuration choices and distribution shift, represented by the Hellinger radius~$\rho$ in Equation 6. Each point corresponds to a specific unlearning configuration applied to one of the three LLMs, with color indicating the resulting unified risk value. On horizontal axis, configurations vary in aggressiveness, while the $\rho$-axis reflects increasing deviation between the reference distribution and a shifted test distribution. The blue translucent plane denotes a chosen risk threshold~$\hat{\alpha}_{\rho}$, allowing us to identify configurations that remain acceptable under a given level of distribution shift. Points lying below the plane correspond to configurations that satisfy the risk requirement, while those above violate it. The figure highlights several trends: unified risk grows as the Hellinger radius increases, indicating reduced robustness under larger distribution drift, and different model families occupy distinct regions of the risk landscape, revealing model-dependent sensitivity to configuration settings. This joint visualization demonstrates how FROC enables principled assessment of configuration validity, providing a unified view of unlearning behavior across models, configurations, and robustness levels.


\subsubsection{\textbf{Impact of Reference Set Size on Conformal Unlearning Risk}}


\begin{figure}[t]
\centering
\includegraphics[width=0.6\linewidth]{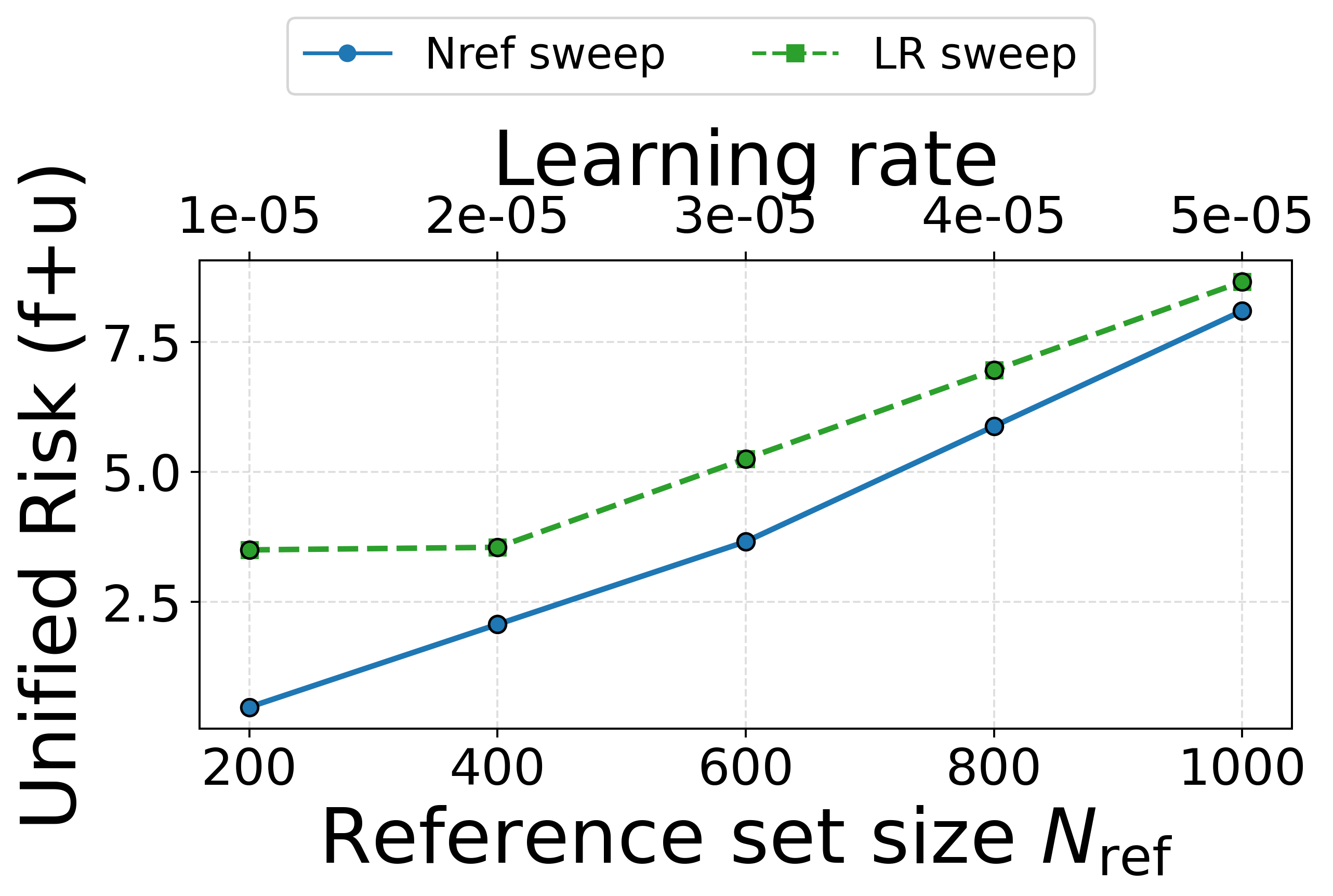}
\caption{Effect of reference set size $N_{ref}$ on unified risk $R$, run on LLAMA3.1-8B.}
\label{fig:nref_effect}
\end{figure}

The results also demonstrate that a larger reference set size \( N_{\text{ref}} \) and more robust unlearning configurations \( \lambda \), such as an increased number of ascent steps, which significantly contribute to reducing conformal unlearning risk. Figure \ref{fig:nref_effect} illustrates how the unified risk $R$ changes as two key unlearning control factors, reference-set size $N_{\mathrm{ref}}$ and learning rate are independently varied for LLaMA3.1-8B. The blue curve shows that increasing $N_{\mathrm{ref}}$ leads to higher unified risk, reflecting the fact that a larger reference set provides a stricter evaluation of both forgetting and utility preservation. As $N_{\mathrm{ref}}$ grows, deviations in the forget and retain behaviors are more easily detected, thereby increasing the overall risk score. The green dashed curve depicts the effect of modifying the learning rate used during unlearning updates. Higher learning rates intensify the forgetting signal but also introduce larger perturbations to the retain distribution, causing the unified risk to rise more sharply. Taken together, the two curves demonstrate that $R$ responds consistently to both evaluation strictness and update aggressiveness: unified risk increases when (i) a larger reference set enforces a tighter assessment of model behavior, or (ii) the unlearning step size becomes more disruptive. This visualization shows how FROC enables systematic exploration of unlearning configurations and offers an interpretable measure for choosing stable operating regimes.

\subsubsection{\textbf{Impact of Unlearning Configuration Set on Conformal Unlearning Risk}}

\begin{figure}[t]
\centering
\includegraphics[width=0.7\linewidth]{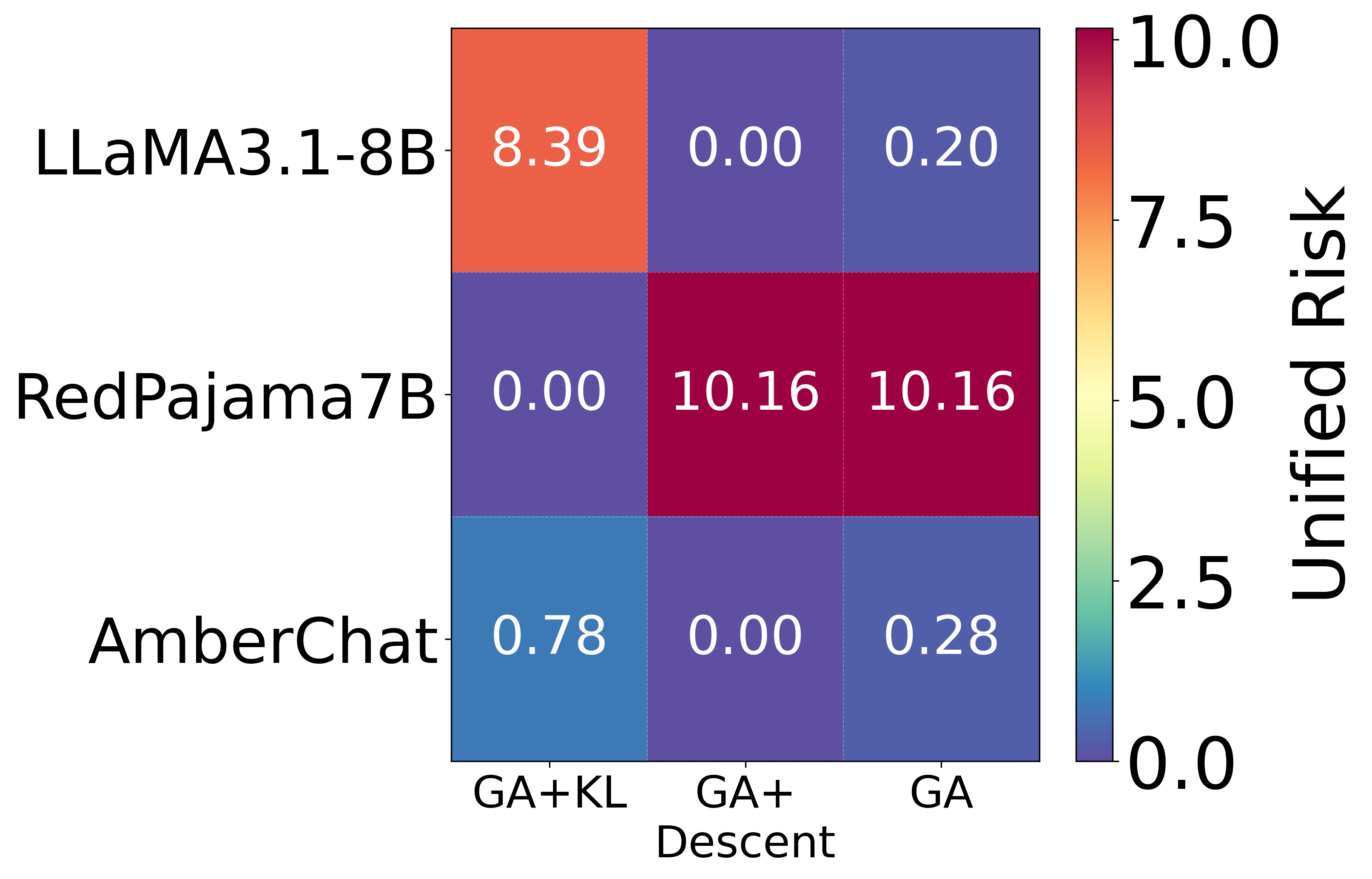}
\caption{Heatmap of Risk-Controlled Unified Risk Across Models and Methods.}
\label{fig:llm_risk_heatmap}
\end{figure}


The unified-risk heatmap in Figure \ref{fig:llm_risk_heatmap} reveals significant model-specific differences in the stability of unlearning updates. Unified risk combines forgetting sufficiency and utility preservation into a single penalty, with low values indicating safe unlearning and high values signaling harmful drift in the retain distribution. All models and methods share the same configuration. Notably, no method is universally optimal; LLaMA3.1-8B and AmberChat favor GA+Descent, achieving near-zero risk, while RedPajama-7B shows GA+KL as the only low-risk configuration. This inversion highlights that unlearning safety is closely linked to a model’s pretraining geometry and inductive biases, emphasizing the need for model-adaptive unlearning and the utility of unified risk as a metric for stable configurations across architectures.

\section{Conclusion}


Our work introduces FROC, a unified framework for risk-optimized control in machine unlearning. FROC provides a continuous, model-agnostic measure of unlearning quality by integrating forgetting sufficiency and utility preservation into a single risk function. It uses a conformal-style formulation with the Hellinger radius for configuration assessment. Results across multiple LLM architectures show that unified risk captures meaningful differences between unlearning configurations, reveals trade-offs driven by hyperparameters like reference set size and learning rate, and offers risk control for stable operating regions. These findings highlight FROC's value as an evaluation tool for unlearning pipelines. 
Future work includes enhancing stability of FROC by tolerating distribution shifts, and in a longer run, achieving provable guarantee of the evaluation results.

\section*{Acknowledgment}
This research / project is supported by the National Research Foundation, Singapore and Infocomm Media Development Authority under its Trust Tech Funding Initiative. Any opinions, findings and conclusions or recommendations expressed in this material are those of the author(s) and do not reflect the views of National Research Foundation, Singapore and Infocomm Media Development Authority.


\bibliographystyle{IEEEtran}
\bibliography{ref}

\end{document}